\documentclass{article}

\usepackage{arxiv}
\usepackage[utf8]{inputenc} 
\usepackage[T1]{fontenc}    
\usepackage{hyperref}       
\usepackage{url}            
\usepackage{booktabs}       
\usepackage{amsfonts}       
\usepackage{nicefrac}       
\usepackage{microtype}      
\usepackage{lipsum}

\usepackage{float}
\usepackage{graphicx}
\usepackage{multirow}

\title{Brazilian Lyrics-Based Music Genre Classification Using a BLSTM Network}

\author{
  Raul~de~Araújo~Lima \\
  Department of Informatics, PUC-Rio\\
  Rua Marques de Sao Vicente 225 \\
  Rio de Janeiro, RJ, 22420-030, Brazil \\
  \texttt{rlima@inf.puc-rio.br} \\
   \And
 Rômulo~César~C.~de~Sousa \\
  Department of Informatics, PUC-Rio\\
  Rua Marques de Sao Vicente 225 \\
  Rio de Janeiro, RJ, 22420-030, Brazil \\
  \texttt{rsousa@inf.puc-rio.br} \\
    \And
 Simone~Diniz~Junqueira~Barbosa \\
  Department of Informatics, PUC-Rio\\
  Rua Marques de Sao Vicente 225 \\
  Rio de Janeiro, RJ, 22420-030, Brazil \\
  \texttt{simone@inf.puc-rio.br} \\
    \And
 Hélio~Cortês~Vieira~Lopes \\
  Department of Informatics, PUC-Rio\\
  Rua Marques de Sao Vicente 225 \\
  Rio de Janeiro, RJ, 22420-030, Brazil \\
  \texttt{lopes@inf.puc-rio.br} \\
}

\begin{document}
\maketitle

\begin{abstract}
Organize songs, albums, and artists in groups with shared similarity could be done with the help of genre labels. In this paper, we present a novel approach for automatic classifying musical genre in Brazilian music using only the song lyrics. This kind of classification remains a challenge in the field of Natural Language Processing. We construct a dataset of $138,368$ Brazilian song lyrics distributed in 14 genres. We apply SVM, Random Forest and a Bidirectional Long Short-Term Memory (BLSTM) network combined with different word embeddings techniques to address this classification task. Our experiments show that the BLSTM method outperforms the other models with an F1-score average of $0.48$. Some genres like \textit{gospel}, \textit{funk-carioca} and \textit{sertanejo}, which obtained $0.89$, $0.70$ and $0.69$ of F1-score, respectively, can be defined as the most distinct and easy to classify in the Brazilian musical genres context.
\end{abstract}


\section{Introduction}

Music is part of the day-to-day life of a huge number of people, and many works try to understand the best way to classify, recommend, and identify similarities between songs. Among the tasks that involve music classification, genre classification has been studied widely in recent years~\cite{ying2012genre} since musical genres are the main top-level descriptors used by music dealers and librarians to organize their music collections~\cite{scaringella2006automatic}.

Automatic music genre classification based only on the lyrics is considered a challenging task in the field of Natural Language Processing (NLP). Music genres remain a poorly defined concept, and boundaries between genres still remain fuzzy, which makes the automatic classification problem a nontrivial task~\cite{scaringella2006automatic}. 

Traditional approaches in text classification have applied algorithms such as Support Vector Machine (SVM) and Na\"ive Bayes, combined with handcraft features (POS and chunk tags) and word count-based representations, like bag-of-words.  More recently, the usage of Deep Learning methods such as Recurrent Neural Networks (RNNs) and Convolutional Neural Networks (CNNs) has produced great results in text classification tasks.

Some works like \cite{laurier2008multimodal}, \cite{hu2009lyric} \cite{hu2010lyrics} focus on classification of mood or sentiment of music based on its lyrics or audio content. Other works, like \cite{scaringella2006automatic}, and \cite{tsaptsinos2017lyrics}, on the other hand, try to automatically classify the music genre; and the work \cite{fell2014lyrics} tries to classify, besides the music genre, the best and the worst songs, and determine the approximate publication time of a song. 

In this work, we collected a set of about 130 thousand Brazilian songs distributed in 14 genres. We use a Bidirectional Long Short-Term Memory (BLSTM) network to make a lyrics-based music genre classification. We did not apply an elaborate set of handcraft textual features, instead, we represent the lyrics songs with a pre-trained word embeddings model, obtaining an F1 average score of $0.48$. Our experiments and results show some real aspects that exist among the Brazilian music genres and also show the usefulness of the dataset we have built for future works.

This paper is organized as follows. In the next section, we cite and comment on some related works. Section~\ref{sec:methods} describes our experiments from data collection to the proposed model, presenting some important concepts. Our experimental results are presented in Section~\ref{sec:results}, and Section~\ref{sec:conclusion} presents our concluding remarks and future work.

\section{Related Works}
\label{sec:related_works}

Several works have been carried out to add textual information to genre and mood classification. Fell and Sporleder~\cite{fell2014lyrics} used several handcraft features, such as vocabulary, style, semantics, orientation towards the world, and song structure to obtain performance gains on three different classification tasks: detecting genre, distinguishing the best and the worst songs, and determining the approximate publication time of a song. The experiments in genre classification focused on eight genres: Blues, Rap, Metal, Folk, R\&B, Reggae, Country, and Religious. Only lyrics in English were included and they used an SVM with the default settings for the classification. 

Ying \textit{et al.}~\cite{ying2012genre} used Part-of-Speech (POS) features extracted from lyrics and combined them with three different machine learning techniques -- k-Nearest-Neighbor, Na\"ive Bayes, and Support Vector Machines -- to classify a collection of 600 English songs by the genre and mood. 

Zaanen and Kanters~\cite{van2010automatic} used the term frequency and inverse document frequency statistical metrics as features to solve music mood classification, obtaining an accuracy of more than 70\%.

In recent years, deep learning techniques have also been applied to music genre classification. This kind of approach typically does not rely on handcraft features or external data. In~\cite{tsaptsinos2017lyrics}, the authors used a hierarchical attention network to perform the task in a large dataset of nearly half a million song lyrics, obtaining an accuracy of more than 45\%. Some papers such as~\cite{kumar2018genre} used word embedding techniques to represent words from the lyrics and then classify them by the genre using a 3-layer Deep Learning model.

\section{Methods}
\label{sec:methods}

In this chapter we present all the major steps we have taken, from obtaining the dataset to the proposed approach to address the automatic music genre classification problem.

\subsection{Data Acquisition}

In order to obtain a large number of Brazilian music lyrics, we created a crawler to navigate into the \textit{Vagalume}\footnote{https://www.vagalume.com.br/} website, extracting, for each musical genre, all the songs by all the listed authors. The implementation of a crawler was necessary because, although the \textit{Vagalume} site provides an API, it is only for consultation and does not allow obtaining large amounts of data. The crawler was implemented using Scrapy\footnote{https://scrapy.org/}, an open-source and collaborative Python library to extract data from websites.

\begin{figure}[H]
\centering
\includegraphics[width=0.8\linewidth]{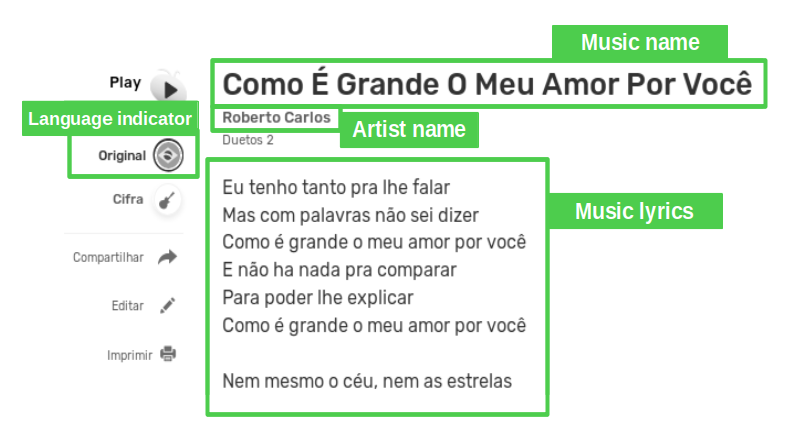}
\caption{An example of a Vagalume's song web page}
\label{fig:vagalume_music}
\end{figure}

From the Vagalume's music web page, we collect the song title and lyrics, and the artist name. The genre was collected from the page of styles\footnote{\url{https://www.vagalume.com.br/browse/style/}}, which lists all the musical genres and, for each one, all the artists. We selected only 14 genres that we consider as representative Brazilian music, shown in Table~\ref{tab:dataset_info}. Figure~\ref{fig:vagalume_music} presents an example of the Vagalume's music Web page with the song \textit{``Como é grande o meu amor por você''}\footnote{\url{https://www.vagalume.com.br/roberto-carlos/como-e-grande-o-meu-amor-por-voce-letras.html}}, of the Brazilian singer Roberto Carlos. Green boxes indicate information about music that can be extracted directly from the web page. From this information, the language in which the lyrics are available can be obtained by looking at the icon indicating the flag of Brazil preceded by the \textit{``Original''} word.

After extracting data, we obtained a set of $138,368$ songs distributed across 14 genres. Table~\ref{tab:dataset_info} presents the number of songs and artists by genre. In order to use the data to learn how to automatically classify genre, we split the dataset into tree partitions: training ($96,857$ samples), validation ($27,673$ samples), and test ($13,838$ samples). The total dataset and splits are available for download\footnote{\url{https://drive.google.com/open?id=1b681ChByK737CpASYImFB4GfuqPdrvBN}}.

\begin{table}[H]
\centering
\caption{The number of songs and artists by genre}
\label{tab:dataset_info}
\begin{tabular}{@{}lrr@{}}
\hline
\textbf{Genre} & \textbf{\#songs} & \textbf{\#artists} \\ \hline
Gospel         & 33,344            & 800                \\
Sertanejo      & 27,417            & 543                \\
MPB            & 16,785            & 282                \\
Forr\'o         & 11,861            & 191                \\
Pagode         & 8,199             & 174                \\
Rock           & 8,188             & 396                \\
Samba          & 6,221             & 111                \\
Pop            & 4,629             & 338                \\
Ax\'e            & 4,592             & 63                 \\
Funk-carioca   & 4,557             & 279                \\
Infantil       & 4,550             & 70                 \\
Velha-guarda   & 3,179             & 24                 \\
Bossa-nova     & 3,105             & 38                 \\
Jovem-guarda   & 1,741             & 18                 \\ \hline
\end{tabular}
\end{table}

\subsection{Word Embeddings}

Word embeddings is a technique to represent words as real vectors, so that these vectors maintain some semantic aspects of the real words. Basically, vectors are computed by calculating probabilities of the context of words, with the intuition that semantically similar words have similar contexts, and must therefore have similar vectors.

Word2Vec\footnote{\url{https://code.google.com/archive/p/word2vec/}}, by Mikolov \textit{et al.}~\cite{mikolov2013distributed}, is one of the first and most widely used algorithms to make word embeddings. It has two architectures to compute word vectors: Continuous Bag-Of-Words (CBOW) and Skip-gram. CBOW gets a context as input and predicts the current word, while Skip-gram gets the current word as input and predicts its context. 

In this work, we use the Python Word2Vec implementation provided by the Gensim\footnote{https://radimrehurek.com/gensim/index.html} library. The Portuguese pre-trained word embeddings created by~\cite{hartmann2017portuguese} and available for download\footnote{\url{http://nilc.icmc.usp.br/embeddings}} was used to represent words as vectors. We only used models of dimension 300 and, for Word2Vec, Wang2Vec, and FastText, skip-gram architectured models.

\subsection{Bidirectional Long Short-Term Memory}

\begin{figure}[htb]
\centering
\includegraphics[width=.7\textwidth]{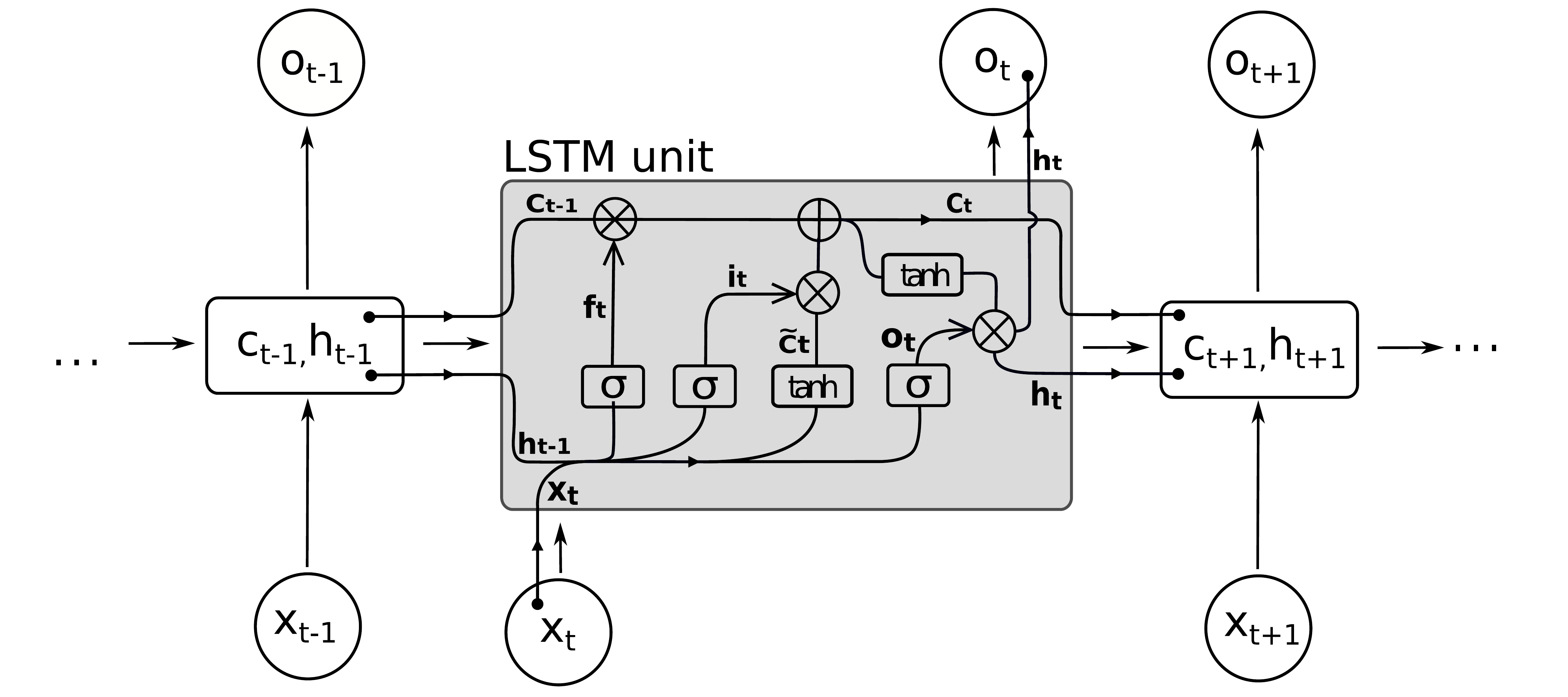}
\caption{The Long Short-Term Memory unit.}
\label{fig:lstm_unit}
\end{figure}

Long Short-Term Memory (LSTM) is a specification of Recurrent Neural Network (RNN) that was proposed by Hochreiter and Schmidhuber~\cite{hochreiter1997long}. This kind of network is widely used to solve classification of sequential data and is designed to capture time dynamics through graph cycles. Figure~\ref{fig:lstm_unit} presents an LSTM unity, which receives an input from the previous unit, processes it, and passes it to the next unit.

The following equations are used to update $C_t$ and $h_t$ values.
$$f_t = \sigma(W_f h_{t-1} + U_f x_t + b_f)$$
$$i_t = \sigma(W_i h_{t-1} + U_i x_t + b_i)$$
$$\widetilde{C}_t = tanh(W_C h_{t-1} + U_C x_t + b_C)$$
$$C_t = f_t \times C_{t-1} + i_t \times \widetilde{C}_t$$
$$o_t = \sigma(W_o h_{t-1} + U_o x_t + b_o)$$
$$h_t = o_t \times tanh(C_t)$$
where $W_f$, $W_i$, $W_C$, $W_o$ are the weight matrices for $h_{t-1}$ input; $U_f$, $U_i$, $U_C$, $U_o$ are the weight matrices for $x_t$ input; and $b_f$, $b_i$, $b_C$, $b_o$ are the bias vectors.

Basically, a Bidirectional LSTM network consists of using two LSTM networks: a forward LSTM and a backward LSTM. The intuition behind it is that, in some types of problems, past and future information captured by forward and backward LSTM layers are useful to predict the current data.

\subsection{Proposed Approach}

\begin{figure}[htb]
\centering
\includegraphics[width=0.6\textwidth]{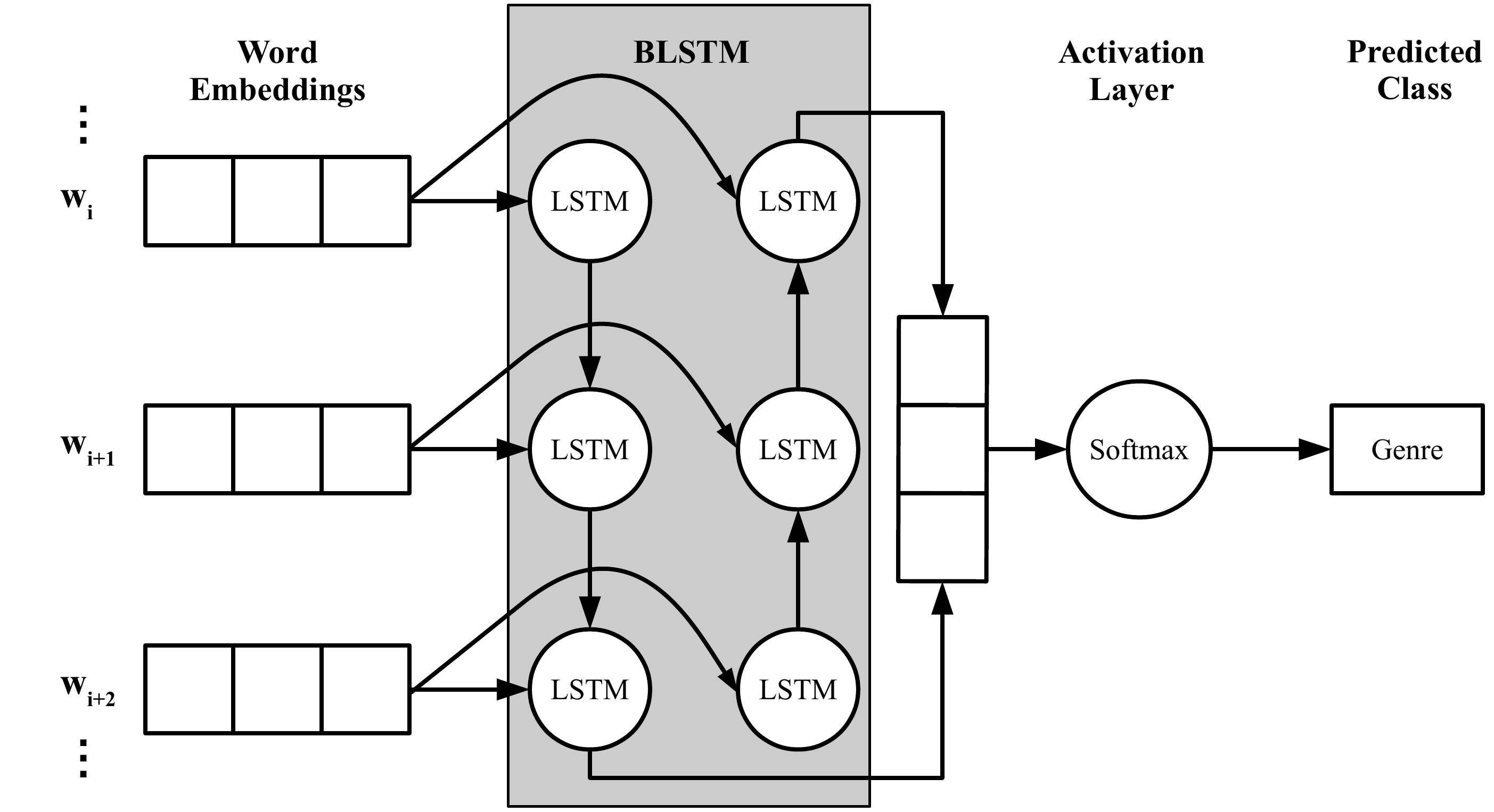}
\caption{Our BLSTM model architecture}
\label{fig:blstm}
\end{figure}

Our proposed approach consists of three main steps. Firstly, we concatenate the title of the song with its lyrics, put all words in lower case and then we clean up the text by removing line breaks, multiple spaces, and some punctuation (,!.?). Secondly, we represent the text as a vector provided by a pre-trained word embeddings model. For classical learning algorithms like SVM and Random Forest, we generate, for each song, a vectorial representation by calculating the average of the vectors of each word in the song lyrics that can be can be expressed by the equation below: $$vector(L) = \frac{1}{n} \sum_{w \in L} vector(w)$$ where $L$ is the song lyrics, $w$ is a word in $L$, and $n$ is the number of words in $L$. If a word does not have a vector representation in the word embeddings model, it is not considered in the equation. For the BLSTM algorithm,  the representation was made in the format of a matrix, as shown in Figure~\ref{fig:blstm}, where each line is a  vector representation of a word in the lyrics. In the third step, we use as features the generated representation for the genre classification tasks using SVM, Random Forests, and BLSTM.

\section{Experimental Results}
\label{sec:results}

In this section, we describe our experiments.  We used the Linear SVM and Random Forest Scikit-learn\footnote{\url{https://scikit-learn.org/}} implementations and Keras\footnote{\url{https://keras.io/}} on top of TensorFlow\footnote{\url{https://www.tensorflow.org/}} for the BLSTM implementation. In this study, we did not focus on finding the best combination of parameters for the algorithms, so that for SVM we used the default parameters, and for Random Forest we used a number of 100 trees. Our BLSTM model was trained using 4 epochs, with Adam optimizer, and 256 as the size of the hidden layer.  

As we can see in Table~\ref{tab:results}, our BLSTM approach outperforms the other models with an F1-score average of $0.48$. In addition, we can note that the use of Wang2Vec pre-trained word embeddings made it possible to obtain better F1-score results in BLSTM, which is not necessarily noticed in other cases, since for SVM and Random Forest, Glove and FastText, respectively, were the techniques that obtained better F1-scores. 

\begin{table}[ht]
\centering
\caption{Classification results for each classifier and word embeddings model combination}
\label{tab:results}
\begin{tabular}{llrrr}
\hline
\textbf{System} & \textbf{Model} & \textbf{Precision} & \textbf{Recall} & \textbf{F1-score} \\ \hline
\multirow{4}{*}{SVM} 
 & Word2Vec & 0.290 & 0.143 & 0.120 \\
 & Wang2Vec & 0.297 & 0.144 & 0.121 \\
 & FastText & 0.278 & 0.144 & 0.120 \\
 & Glove & 0.296 & 0.144 & \textbf{0.124} \\ \hline
\multirow{4}{*}{Random Forest} 
 & Word2Vec & 0.388 & 0.197 & 0.212 \\
 & Wang2Vec & 0.386 & 0.197 & 0.207  \\
 & FastText & 0.404 & 0.203 & \textbf{0.215} \\
 & Glove & 0.394 & 0.199 & 0.210 \\ \hline
\multirow{4}{*}{BLSTM} 
 & Word2Vec & 0.492 & 0.454 & 0.465 \\
 & Wang2Vec & \textbf{0.515} & \textbf{0.460} &  \textbf{0.477} \\
 & FastText & 0.417 & 0.350 &  0.360 \\
 & Glove    & 0.492 & 0.460 &  0.470 \\ \hline
\end{tabular}
\end{table}

Table~\ref{tab:results_BLSTM} shows the BLSTM classification results for each genre. We can see that the genres \textit{gospel}, \textit{funk-carioca} and \textit{sertanejo} have a greater distinction in relation to the other genres, since they were better classified by the model. In particular, \textit{funk-carioca} obtained a good classification result although it did not have a large number of collected song lyrics.

In gospel song lyrics, we can identify some typical words, such as \textit{``Deus''} (God) , \textit{``Senhor''} (Lord), and \textit{``Jesus''} (Jesus); in \textit{funk-carioca}, songs  have the words \textit{``bonde''} (tram), \textit{``ch\~ao''} (floor) and \textit{``baile''} (dance ball), all used as slang; in \textit{sertanejo}, some of the most common words are \textit{``amor''} (love), \textit{``cora\c{c}\~ao''} (heart) and \textit{``saudade''} (longing). The occurrence of these typical words could contribute to the higher performance of F1-scores in these genres.

\begin{table}[ht]
\centering
\caption{Detailed result of BLSTM}
\label{tab:results_BLSTM}
\begin{tabular}{lr}
\hline
\textbf{Genre} & \textbf{F1-score} \\ \hline
Gospel & 0.89 \\
Funk-carioca & 0.70 \\
Sertanejo & 0.69 \\
Forr\'o & 0.53 \\
Ax\'e & 0.49 \\
MPB & 0.49 \\
Pagode & 0.48 \\
Infantil & 0.47 \\
Rock & 0.46 \\
Velha-guarda & 0.38 \\
Samba & 0.35 \\
Bossa-nova & 0.31 \\
Pop & 0.26 \\
Jovem-guarda & 0.19 \\ \hline
\textit{Average} & \textit{0.481} \\ \hline
\end{tabular}
\end{table}

The \textit{bossa-nova} and \textit{jovem-guarda} genres, which have few instances in the dataset, are among the most difficult ones to classify using the model. The \textit{pop} genre, by contrast, has a small distribution between the number of songs and the number of artists, and could not be well classified by our model. This may indicate that our model was unable to identify a pattern due to the low number of songs per artist, or that the song lyrics of this genre cover several subjects that are confused with other genres.

\begin{figure}[ht]
\centering
\includegraphics[width=0.7\textwidth]{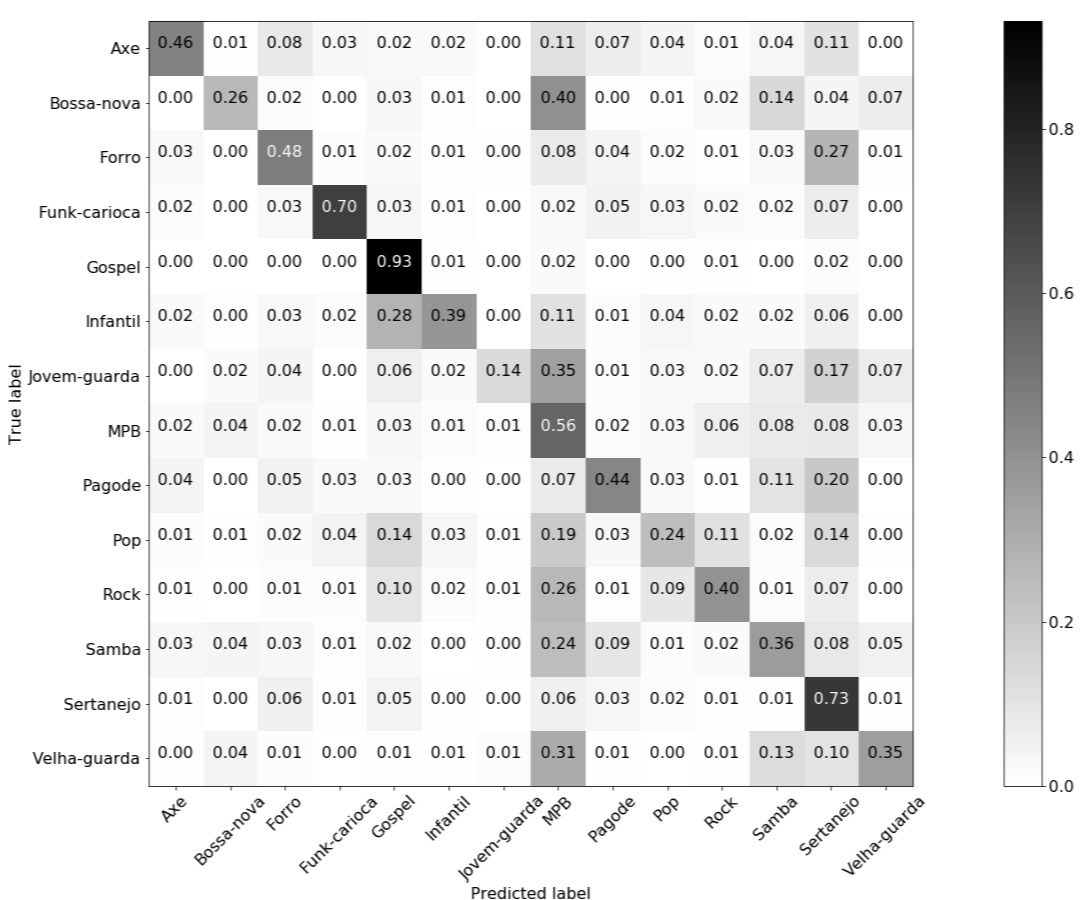}
\caption{Normalized confusion matrix}
\label{fig:cm}
\end{figure}

Figure \ref{fig:cm} shows the confusion matrix of the results produced by our BLSTM model. We can notice that many instances of class  \textit{forr\'o} are often confused with class \textit{sertanejo}. Indeed, these two genres are very close. Both \textit{Forr\'o} and \textit{sertanejo} have as theme the cultural and daily aspects of the Northeast region of Brazil. Instances of class \textit{infantil} are often confused with class \textit{gospel}: in  \textit{infantil} we have music for children for both entertainment and education. In some of the songs, songwriters try to address religious education, which could explain the confusion between those genres. The \textit{MPB} (Brazilian Popular Music) genre was the most confused of all, which may indicate that song lyrics of this genre cover a wide range of subjects that intersect with other genres.

\section{Conclusion and Future Works}
\label{sec:conclusion}

In this work we constructed a dataset of $138,368$ Brazilian song lyrics distributed in 14 genres. We applied SVM, Random Forest, and a Bidirectional Long Short-Term Memory (BLSTM) network combined with different word embeddings techniques to address the automatic genre classification task based only on the song lyrics. We compared the results between the different combinations of classifiers and word embedding techniques, concluding that our BLSTM combined with the Wang2Vec pre-trained model obtained the best F1-score classification result. Beside the dataset construction and the comparison of tools, this work also evidences the lack of an absolute superiority between the different techniques of word embeddings, since their use and efficiency in this specific task showed to be very closely related to the classification technique.

As future work, it is possible to explore the dataset to identify genre or artist similarities, generating visualizations that may or may not confirm aspects pre-conceived by the consumers of Brazilian music. It is also possible to perform classification tasks by artists of a specific genre.

\bibliographystyle{unsrt}  
\bibliography{references}  


\end{document}